\documentclass{article}

\usepackage{arxiv}

\usepackage[utf8]{inputenc} 
\usepackage[T1]{fontenc}    
\usepackage{hyperref}       
\usepackage{url}            
\usepackage{booktabs}       
\usepackage{amsfonts}       
\usepackage{nicefrac}       
\usepackage{microtype}      
\usepackage{lipsum}
\usepackage{graphicx}
\usepackage{subfig}

\title{Control of Biohybrid Actuators using NeuroEvolution}

\author{
    Hugo Alcaraz-Herrera\\
    Unconventional Computing Laboratory, \\
    College of Arts, Technology and Environment, \\
    University of the West of England, \\
    Bristol, BS16 1QY, United Kingdom \\ \texttt{hugo.alcaraz@uwe.ac.uk}
\And
    Michail-Antisthenis Tsompanas  \\
    Unconventional Computing Laboratory \& \\
    School of Computing \& Creative Technologies,\\
    College of Arts, Technology and Environment, \\
    University of the West of England,\\ 
    Bristol, BS16 1QY, United Kingdom \\
    \texttt{antisthenis.tsompanas@uwe.ac.uk}
\And
       Andrew Adamatzky \\
       Unconventional Computing Laboratory,\\
    College of Arts, Technology and Environment, \\
    University of the West of England,\\ 
    Bristol, BS16 1QY, United Kingdom \\
\And
       Igor Balaz\\
       Laboratory for Meteorology, Physics and Biophysics,\\ Faculty of Agriculture, \\
       University of Novi Sad, \\ Trg Dositeja Obradovica 8, 21000, Novi Sad, Serbia
}

\begin{document}
\maketitle

\begin{abstract}
In medical-related tasks, soft robots can perform better than conventional robots because of their compliant building materials and the movements they are able perform. However, designing soft robot controllers is not an easy task, due to the non-linear properties of their materials. Since human expertise to design such controllers is yet not sufficiently effective, a formal design process is needed. The present research proposes neuroevolution-based algorithms as the core mechanism to automatically generate controllers for biohybrid actuators that can be used on future medical devices, such as a catheter that will deliver drugs. The controllers generated by methodologies based on Neuroevolution of Augmenting Topologies (NEAT) and Hypercube-based NEAT (HyperNEAT) are compared against the ones generated by a standard genetic algorithm (SGA). In specific, the metrics considered are the maximum displacement in upward bending movement and the robustness to control different biohybrid actuator morphologies without redesigning the control strategy. Results indicate that the neuroevolution-based algorithms produce better suited controllers than the SGA. In particular, NEAT designed the best controllers, achieving up to 25\% higher displacement when compared with SGA-produced specialised controllers trained over a single morphology and 23\% when compared with general purpose controllers trained over a set of morphologies.
\end{abstract}


\section{Introduction}

Soft robotics is a sub-field of robotics that studies machines built with flexible and ductile materials, such as silicone rubbers \cite{Rus2015}. This type of robots have demonstrated better performance in specific tasks related to 
healthcare \cite{Hsiao2019}, due to their morphology and behaviour that are heavily inspired by living organisms. 

Despite their promising applicability, soft robots face significant challenges, 
i.e., defining an adequate morphology design. Under a traditional approach of robot designing, considerable time and material resources are utilised since numerous alternative prototypes are tested physically \cite{Schulz2016}. On the other hand, under a soft robots approach, designing process is more complex due to the materials' flexibility and mechanical properties being non-linear and difficult to characterise \cite{Hiller2014}.

When a suitable design is found, the next step consists of designing an appropriate controller for the soft robot, a process that can be considered as intricate as finding a suitable morphology. The required strategies to control soft robots have two main considerations: (i) the soft materials constituting the robot can deform at every point, resulting in infinite degrees of freedom, including bending, extension, contraction, and torsion, and (ii) soft materials present non-linear and time-dependent properties. These aspects make modeling of soft robot behaviour and movement a difficult task \cite{Wang2022}.

A methodology that can assist the design of controllers for soft robots which is worth investigating is Neuroevolution (NE). NE focuses on evolving the topology and weights of artificial neural networks (ANNs) employing a genetic algorithm (GA) methodology. Arguably, the most efficient NE algorithm has proved to be Neuroevolution of Augmenting Topologies (NEAT) \cite{Stanley2002}. Furthermore, under the rationale that natural structures are composed of shape repetition and patters, an extension of NEAT was developed, namely Hypercube-based Neuroevolution of Augmenting Topologies (HyperNEAT) \cite{Stanley2009}. HyperNEAT evolves a singular type of ANNs named Compositional Pattern-Producing Networks (CPPNs), whose difference to traditional ANNs focuses around the use of periodic functions (e.g., sine and square wave), in order to generate patterns, like symmetry and repetition that help evolve more interesting topologies \cite{Stanley2007cppn}.

The fundamental objective of the presented research is assessing the suitability of NEAT and HyperNEAT as design engines for controllers of {\em biohybrid actuators} (BHAs), a particular type of soft robots' components that are built utilising biological material, such as tissues or cells. Specifically, we analyse the capabilities of the NEAT and HyperNEAT to design controllers whose objective is to induce an upward bending movement to a given BHA. That BHA may be embodied into a catheter for targeted drug delivery to areas of the human body that are difficult to reach.





\section{{Background}}\label{sec:background}

NEAT has been previously used as a controller design engine in robotic applications. For instance, NEAT is implemented to design controllers for a robot in a crowded environment \cite{Seriani2021}. A ray-casting model can be implemented in industrial robots with relative acts as the robot's perception, which can perceive objects at the current time and has a memory of previously perceived objects. Controllers are first assessed in simulated environments and subsequently in a minimal physical implementation. Results advocate that controllers generated by NEAT converge to a suitable design in both environments. 

In another exemplar research, NEAT is applied to design locomotive controllers \cite{Tibermacine2014}. The scope of the study focuses on virtual creatures that are simulated in a physics engine called {\em Open, Dynamic Engine} (ODE). The controllers generated by NEAT are compared against controllers evolved in a more standard approach 
and results indicate that controllers designed by NEAT exhibit better performance since they consider the locomotive features of morphologies and tend to be more robust than those generated by the traditional approach.



Finally, NEAT has also been utilised to design controllers for autonomous vehicles. For instance, a reactive navigation hybrid controller for non-holonomic mobile robots was studied \cite{Caceres2017}. Experiments were conducted in a simulation platform developed by the authors and are based on the kinematic model of a car-like robot. Results suggest that controllers designed by NEAT exhibited the expected performance of controlling the kinematics of car-like robots in an unknown environment, avoiding obstacles and reaching the target point.

Moreover, HyperNEAT has also been utilised to design robot controllers, as well as their shapes. For instance, a method that aims to find suitable morphologies and controllers has been introduced \cite{Tanaka2022}. The method was tested in four scenarios considering the adaptation level observed in the robots used for experimentation. The results suggested that the approaches can generate morphologies and that their controllers are capable of suitably performing on the given tasks. 

Furthermore, HyperNEAT has been used to design locomotion controllers utilized by autonomous crawler robots with flippers \cite{Sokolov2017}. In order to evaluate their performance, controllers were tested in the ROS/Gazebo simulator. The inputs of CPPNs were Light Detection and Ranging (LIDAR) data, the robot's position, orientation, flipper angle, and track velocities. The outputs were commands to control the flipper angles and track velocities. The results indicated that HyperNEAT-based controllers help crawler robots navigate and overcome obstacles in three-dimensional environments.



Finally, an implementation of HyperNEAT to design controllers for quadruped robots has been proposed \cite{Risi2013}. The model had as input the morphologies of robots. Moreover, the output was neural-network-based controllers capable of working with different robot morphologies. The performance of controllers was assessed utilising three morphologies, and they were compared against static controllers. Results suggested that HyperNEAT identified the relationship between morphologies and controller architectures, which help to generate suitable performance.

Inspired by the aforementioned works of employing NE for controlling different types of robots (i.e., crawlers, runners and even car-like) to achieve higher displacement, the same methodology is tested here on robots limited to angular-only movement. To the best of the authors' knowledge, this aspect has not been tested before, and insights gained by this research will assist the design methodology of robotic controllers for tasks requiring precision movements, i.e. medical applications.


\section{{Neuroevolution}}\label{sec:neuroevolution}


Two of the most popular NE-based approaches are utilised in this research. Namely, NEAT \cite{Stanley2002}, and an extension of NEAT: HyperNEAT \cite{Stanley2009}.

\subsection{NEAT}\label{sec:neuroevolution_neat}

A well-established method to train ANNs is backpropagation of errors, whereas an alternative method is utilizing evolutionary optimization methods. The fitness functions of such methods can be the minimization of the output errors of ANNs. Namely, NE iteratively applies selection, crossover and mutation operators on a population of randomly initialized networks to discover the best performing one, after a given amount of generations. NEAT algorithm was conceived to alleviate three main ``pathologies'' observed in previous NE algorithms \cite{Stanley2002}. Namely, (i) the lack of appropriate solution representations that would allow the recombination of arbitrary network topologies; (ii) preventing the premature disappearance of novel network topologies discovered during evolution; and (iii) avoiding the use of fitness functions to punish complex topologies of individuals.

\subsection{HyperNEAT}\label{sec:neuroevolution_hyperneat}

A well established extension of NEAT is HyperNEAT. This approach employs NEAT to evolve the topology of a particular type of neural network known by the term Compositional Pattern-Producing Network (CPPN) \cite{Stanley2007}. HyperNEAT evolves CPPNs and takes advantage of their properties to reproduce natural patterns observed in nature, such as symmetry and repetition \cite{Stanley2009}. Two critical differences exist between NEAT and HyperNEAT: (i) Activation functions and (ii) Substrate.


Commonly, NEAT generates ANNs containing hidden nodes that uniformly adapt the sigmoid as their activation function. On the other hand, HyperNEAT can use diverse activation functions in the nodes within CPPNs. Examples of the possible activation functions are trigonometric, periodic, and Gaussian. Using different activation functions during evolution allows for exploring a significantly more extensive search space of network topologies.


Due to the pattern reproduction capacity of CPPNs, HyperNEAT can embody the geometry within the domain of the problem. Consequently, those geometrical aspects are considered to determine the topology of ANNs. The geometric layout where these ANNs are defined and where HyperNEAT operates is called {\em substrate}.

\section{{Experimental setup}}\label{sec:experimental_setup}

The main objective of this research is to study the suitability of NEAT and HyperNEAT in designing controllers for BHAs. The capacity to induce an upward bending movement in BHAs is evaluated as the primary suitability measure of each controller. 

\subsection{Voxelyze}\label{sec:experimental_setup_voxelyze}

BHAs are simulated in a physics engine called {\em Voxelyze}, which simulates the physical response of BHAs under specific conditions, such as gravity and ambient viscosity \cite{Hiller2014}. The source code of Voxelyze is freely available and utilized in several optimization studies \cite{Kriegman2020code,Tsompanas2024}. 

In order to evaluate the controllers designed by the proposed methodologies, for the application of moving a catheter, one end of the BHA is considered fixed in place, whereas the other is free to move under the dictated expansion of the active material in its morphology. Thus, to trace the position of the free end of BHAs in the $x,y,z$ axes during time $t$, Voxelyze has been modified to represent BHAs with only one degree of freedom in the $yz$ plane (i.e., they only move vertically). Furthermore, based on the findings of previous research \cite{Tsompanas2024} and laboratory constraints, BHAs are considered within a passive enclosure. 

\begin{figure}[tb!]
  \centering
     \includegraphics[width=0.5\linewidth]{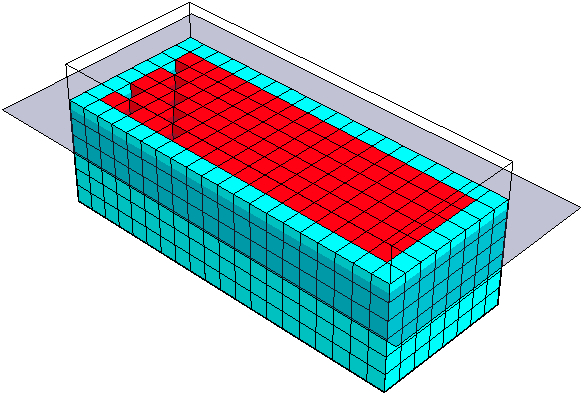}
  \caption{Example of an BHA simulated by Voxelyze.}
  \label{fig:experimental_setup_voxelyze_bha}
\end{figure}

The output of Voxelyze (i.e. the position of the BHA's free tip) is used to evaluate the performance of CPPNs. Based on previous research \cite{Alcaraz2024actuator}, the dimensions of BHA, in terms of voxels, are twenty units in the $x$ axis and eight units in the $y$ and $z$ axes. Figure~\ref{fig:experimental_setup_voxelyze_bha} presents an example of a BHA being simulated by the modified version of Voxelyze.

\subsection{Configuration Scheme}\label{sec:experimental_setup_configuration}

For NEAT and HyperNEAT, the population is composed of 50 individuals (i.e., ANNs and CPPNs representing controllers) and each evolutionary trial lasted 200 generations. Moreover, the activation functions utilised for experimentation are: {\em sine}, {\em negative sine}, {\em absolute value}, {\em negative absolute value}, {\em square}, {\em negative square}, {\em squared absolute value}, {\em negative squared absolute value}, {\em sigmoid}, {\em clamped}, {\em cubical}, {\em exponential}, {\em Gaussian}, {\em hat}, {\em identity}, {\em inverse}, {\em logarithmic}, {\em ReLU}, {\em SeLU}, {\em LeLU}, {\em eLU}, {\em softplus}, {\em hyperbolic tangent}. The specific parameters employed for the evolutionary process of CPPNs are presented in Table~\ref{tab:catheter_parameters}.

\begin{table}
\caption{Parameters utilised to evolve CPPNs under NEAT and HyperNEAT.}
\label{tab:catheter_parameters}
 \begin{center}
  \begin{tabular}{ c c } 
   \toprule
   Parameter & Value \\
   \midrule
   compatibility threshold & 3 \\
   compatibility disjoint coefficient & 1.0 \\ 
   compatibility weight coefficient & 0.5 \\
   maximum stagnation & 25 \\
   survival threshold & 0.2 \\ 
   activation function mutate rate & 0.4 \\
   adding/deleting connection rate & 0.2/0.1 \\
   activating/deactivating connection rate & 0.5\\
   adding/deleting node rate & 0.2/0.1 \\
   \bottomrule
  \end{tabular}
 \end{center}
\end{table}

Furthermore, individuals are initialised with the minimal topology possible: no hidden neurons and input neurons fully connected to output neurons (similar to \cite{Tsompanas2024b}).

\subsubsection{NEAT configuration}\label{sec:experimental_setup_configuration_neat}

Due to BHAs being designed in a discrete three-dimensional layout with two types of voxels representing different materials, the input of controllers considers: (i) the coordinates for each point across the layout and (ii) the type of material. By the term voxel, the basic building block in Voxelyze is defined. Each voxel can be either active (blue color in Fig. \ref{fig:experimental_setup_voxelyze_bha}) or contractile (red color). Furthermore, the output of controllers is the phase offset of each voxel across the layout, dictating the delay in the expansion behaviour of active voxels. Thus, under NEAT, CPPNs are queried as follows:

\begin{equation}\label{eq:experimental_setup_configuration_neat} 
    CPPN(x_i,y_i,z_i,m_i) = pho_i
\end{equation}

\noindent
where the $(x_i,y_i,z_i)$ tuple represents the coordinates of the $i$-th point in the three-dimensional layout and $m_i$ represents the material of the appropriate voxel located in the aforementioned point, which is encoded as follows: 0, absence of a voxel; 1 passive voxel; 3 contractile voxel. Regarding $pho_i$, it represents the phase offset of the $i$-th point of the layout. In order to provide a complete sinusoidal-based contraction movement, the output of CPPNs (i.e., controllers) is clamped in the $[-2\pi, 2\pi]$ range.

\subsubsection{HyperNEAT configuration}\label{sec:experimental_setup_configuration_hyperneat}

Under HyperNEAT, the first defining aspect is the substrate (i.e., ANNs), which has four input neurons due to the discrete three-dimensional layout where BHAs are designed and the type of each voxel. Furthermore, one output neuron is needed to provide the phase offset of each voxel across the three-dimensional layout. Equation~\ref{eq:experimental_setup_configuration_hyperneat} describes how substrates are queried:

\begin{equation}\label{eq:experimental_setup_configuration_hyperneat} 
    substrate(x_i,y_i,z_i,m_i) = pho_i
\end{equation}

\noindent
where the $(x_i,y_i,z_i)$ tuple is the coordinates of the $i$-th point in the three-dimensional layout. Moreover, the $m_i$ variable is related to the voxel type, encoded as before. The $pho_i$ variable is defined as in the above. 

\begin{figure}[tb!]
  \centering
     \includegraphics[width=0.5\linewidth]{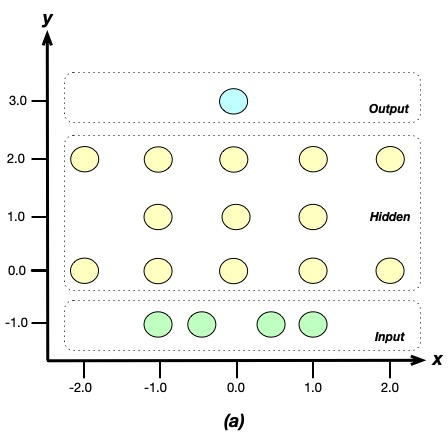}
  \caption{Substrate utilised under HyperNEAT to design BHA controllers.}
  \label{fig:experimental_setup_configuration_hyperneat_substrate}
\end{figure}

The next step consists of allocating the neurons composing the substrate. A series of experiments where the number of hidden layers and the number of neurons per hidden layer is varied in the range $[3,10]$ were conducted to find the optimal allocation. Figure~\ref{fig:experimental_setup_configuration_hyperneat_substrate} depicts the design of the two-dimensional substrate employed throughout the experiments described in this research. Furthermore, the activation function implemented for the neurons composing the substrate is ReLU, since it induces a linear (or close to linear) behaviour and exhibits representational sparsity properties \cite{Glorot2011}.

Due to the evaluation of individuals implying a simulation procedure, the amount of computation time is significant. 
Thus, a client-server implementation was used to take advantage of the distributed computing capacities of this software architecture \cite{Alcaraz2024locomotion}. 



\section{{NEAT vs HyperNEAT}}\label{sec:exp_comparison}

Three metrics are utilised to assess the suitability of NEAT and HyperNEAT to design controllers for BHAs: (i) studying the general performance of controllers in terms of inducing an angular movement on the $yz$ plane to the BHA during a predefined simulated time period; (ii) testing the robustness of controllers; and (iii) analysing the complexity of controllers. Note here that the displacement of BHAs is measured in terms of the length of one voxel.

Moreover, a standard genetic algorithm (SGA) is utilised as a baseline controller design engine. In order to facilitate the implementation of the elements composing the SGA (i.e., individuals, fitness function, and genetic operators), an object-oriented framework was utilised \cite{Alcaraz2022}. Individuals are represented by a bi-dimensional array (i.e., a matrix) of real numbers in the $[-2\pi,2\pi]$ range. 
Regarding the genetic operators, the crossover implementation is {\em two-point} variation with probability of 0.9. 
Regarding mutation, one element of the matrix is randomly chosen and replaced by a random number within the $[-2\pi,2\pi]$ range. This takes place with a probability of 0.1.%

\subsection{General performance}\label{sec:exp_comparison_general}

In this research, the BHAs 
need to bend in a determined direction. The case study presented in this research considers upward bending movement as the target for BHAs controllers. These bending movement is measured utilising the displacement observed in the $yz$ plane. 

\begin{figure*}[tb!]
  \centering
     \includegraphics[width=0.99\linewidth]{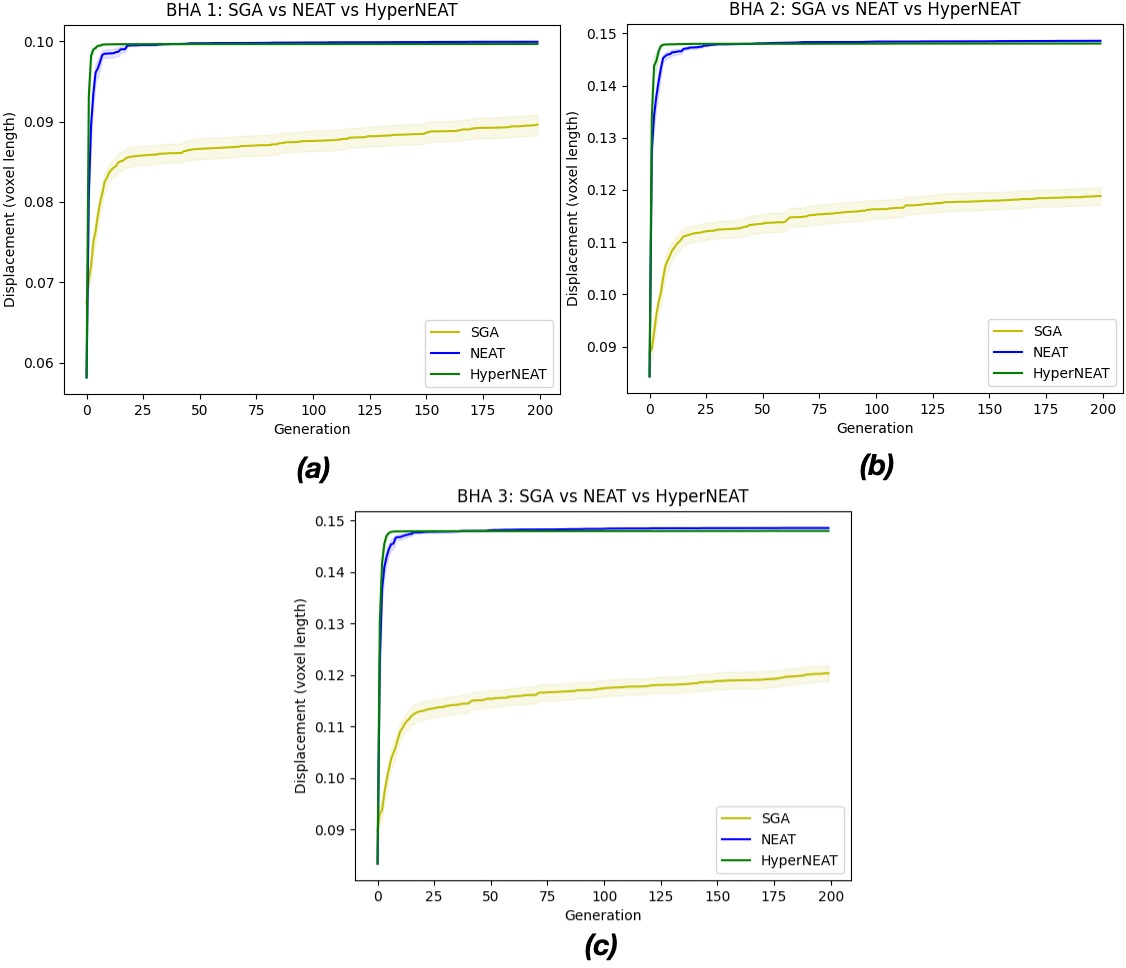}
  \caption{Mean general performance with $\pm 95$ confidence interval (shaded region) under SGA, NEAT, and HyperNEAT using: (a) BHA 1; (b) BHA 2; and (c) BHA 3.}
  \label{fig:exp_comparison_general}
\end{figure*}

Here the general performance of SGA, NEAT, and HyperNEAT in designing controllers capable of inducing upward bending movements to BHAs was analysed. The top three BHA morphologies discovered in previous works \cite{Alcaraz2024actuator} are utilised for experimentation. Figure~\ref{fig:exp_comparison_general} presents the mean performance of the fittest controller with 95\% confidence interval depicted by the shaded regions across 20 evolutionary trials under SGA, NEAT, and HyperNEAT using three different morphologies: (i) BHA 1 (Fig.~\ref{fig:exp_comparison_general}-a); (ii) BHA 2 (Fig.~\ref{fig:exp_comparison_general}-b); and (iii) BHA 3 (Fig.~\ref{fig:exp_comparison_general}-c).

NEAT and HyperNEAT significantly outperform SGA regardless of the BHA morphology utilised. The data collected were tested and proved not to be normally distributed (Shapiro–Wilk test; $p<0.01$). Using the Wilcoxon test, it is feasible to confirm that significant differences in the performance between NE-based approaches and SGA exist (paired Wilcoxon-test; $p<0.01$). Furthermore, NE-based approaches found the fittest controller at the early stages of evolution (i.e., at the first generations of the evolutionary process), whereas SGA tends to gradually evolve at a lower pace. In addition, although the difference in performance between NEAT and HyperNEAT is not clearly visible in all BHAs used, NEAT significantly outperforms HyperNEAT (Shapiro–Wilk test, paired Wilcoxon-test; $p<0.01$). 

Furthermore, the Kruskal-Wallis test is used to confirm significant differences among the performance of all approaches regardless of the BHA  ($p<0.01$). Consequently, is it possible to rank the performance of the three approaches: NEAT $>$ HyperNEAT $>$ SGA (Dunn's test; $p<0.01$).

Results suggest that NE-based approaches can design fitter controllers than those designed by SGA since more elements of the domain problem are considered, namely, the morphology and material of BHAs during evolution. Another relevant aspect that arguably enhances the performance of NEAT and HyperNEAT is the set of properties of CPPNs that help to design controllers that induce upward bending movements that follow a pattern considering the morphology of the BHA.

\subsection{Robustness}\label{sec:exp_comparison_robustness}

A controller can perform adequately inducing upward bending of specific BHA morphologies. However, performance may be different if the BHA morphology changes. This experiment aims to test the robustness of controllers under a more diverse set of morphologies (i.e., how suitable the controller's capacity is to induce upward bending movements regardless of the BHA morphology) designed by SGA, NEAT, and HyperNEAT. For each approach, 20 evolutionary trials were performed. 

Controllers are evaluated using the top nine BHAs discovered in \cite{Alcaraz2024actuator}. Therefore, nine simulations are run utilising Voxelyze, each with a different BHA and its phase offsets generated by the controller. Thus, controllers' robustness (i.e., aptitude) is calculated considering the displacement observed in the $yz$ plane of each simulation. Equation~\ref{eq:exp_comparison_robustness} defines how the aptitude of the controller $c$ is computed:

\begin{equation}\label{eq:exp_comparison_robustness} 
    apt_c= \frac{\sum_{i=1}^{9} displacement_i}{9}
\end{equation}

\begin{figure}[tb!]
  \centering
     \includegraphics[width=0.8\linewidth]{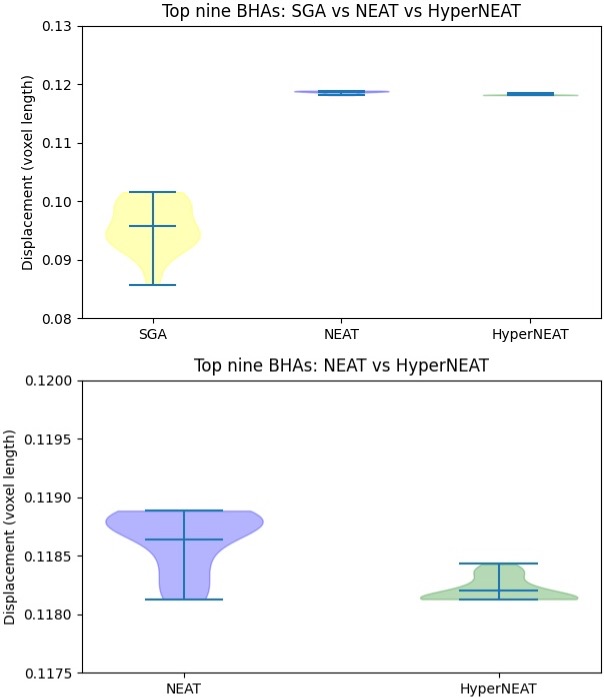}
  \caption{Displacement observed in the $yz$ plane of the top nine BHAs induced by: top - SGA (left), NEAT (centre), and HyperNEAT (right); bottom (close up) - NEAT (left), and HyperNEAT (right).}
  \label{fig:exp_comparison_robustness}
\end{figure}

Figure~\ref{fig:exp_comparison_robustness} depicts violin plots comparing the displacement induced by the fittest controller found by SGA, NEAT and HyperNEAT. Each violin plot presents median, maximum, minimum and kernel density estimation of the frequency distribution of values utilising nine different BHAs.

The displacement induced by all the controllers exhibit significant differences. First, all the data gathered are not normally distributed (Shapiro-Wilk test; $p<0.01$). Then, through the Kruskal-Wallis test, it is feasible to confirm significant differences among the displacement induced by the controllers ($p<0.01$). Consequently, a performance ranking can be defined: NEAT $>$ HyperNEAT $>$ SGA (Dunn's test; $p<0.01$).

Results indicate that despite the number of BHAs involved during the evolutionary process, NE-based approaches are more suitable for designing controllers capable of inducing higher upward bending movements than the standard evolutionary approach. 
Again, NEAT and HyperNEAT are more suitable due to their capacity to consider the morphology and material of BHAs when designing controllers. Furthermore, CPPNs, the core of both techniques, allow the production of patterns that have a significant contribution to the emergence of more efficient movement. Consequently, the upward bending movements induced by NEAT and HyperNEAT controllers follow a pattern based on the morphology of each BHA. 

To better compare the results of the two previous sections, Table \ref{tab:comparison} is provided.

\begin{table}
\caption{Mean displacement (in voxel lengths) achieved by the fittest controller designed by SGA, NEAT and HyperNEAT for one morphology and for a set of 9 morphologies.}
\label{tab:comparison}
 \begin{center}
  \begin{tabular}{ c c c c } 
   \toprule
   Scenario & SGA & NEAT & HyperNEAT\\
   \midrule
   BHA 1 & 0.0896 & 0.0999 & 0.0996 \\
   BHA 2 & 0.1188 & 0.1485 & 0.1480 \\
   BHA 3 & 0.1202 & 0.1485 & 0.1479 \\
   Set of 9 & 0.0957 & 0.1186 & 0.1182 \\
   \bottomrule
  \end{tabular}
 \end{center}
\end{table}

\subsection{Controller complexity}\label{sec:exp_comparison_complexity}

Since the BHAs utilised for experimentation, represent soft robotics components that can be implemented in real life \cite{Alcaraz2024actuator}, the controllers discovered in this study should be feasible to be built. 
Thus, the less complex (i.e., fewer nodes and connections) a controller network required, the easier and more efficient the controller device. 

A crucial aspect to consider for this experiment is that under NEAT, controllers are represented by a CPPNs. In contrast, under HyperNEAT, substrates (i.e. ANNs) represent the controllers. Furthermore, SGA is not considered for this experiment due to the poor performance exhibited previously. Table~\ref{tab:exp_comparison_complexity} shows the mean number of hidden nodes and connections composing the fittest controller across 20 evolutionary trials under NEAT and HyperNEAT. 

\begin{table}
\caption{Mean number of hidden nodes and connections of the fittest controller designed by NEAT and HyperNEAT.}
\label{tab:exp_comparison_complexity}
 \begin{center}
  \begin{tabular}{ c c c } 
   \toprule
   Approach & Hidden Nodes & Connections\\
   \midrule
   NEAT & 2 & 3\\
   HyperNEAT & 13 & 36 \\ 
   \bottomrule
  \end{tabular}
 \end{center}
\end{table}

In general, HyperNEAT produces significantly more complex controllers than NEAT due to the fixed number of hidden neurons of the substrate where HyperNEAT operates. This restricts the exploration of the search space since it only finds the optimal number of connections, their weights, and the bias of neurons. On the other hand, NEAT explores a broader search space that includes the activation functions, the number of neurons (and their bias), and the number of connections (and their weights), arguably allowing it to find more efficient network structures in terms of complexity and performance.  Due to the capacity of NEAT to discover simpler controller networks 
it will be preferred for implementing real-life controllers.


\section{{Conclusions}}\label{sec:conclusions}

This work studies the capacity of NEAT and HyperNEAT to produce suitable controllers for BHAs. Their suitability is compared against a SGA. The performance of the three approaches is analysed under three metrics: (i) general performance (maximum upward bending movement possible to three BHAs); (ii) testing the robustness of controllers produced (maximum upward bending movement on nine BHAs); and (iii) analysing their complexity. For all metrics, 20 evolutionary trials were conducted under the three approaches.

Results suggest that NEAT and HyperNEAT are more suitable for designing controllers for BHAs than SGA, not only for a single morphology but for numerous alternative morphologies. 
In general, NE-based approaches outperform SGA due to: (i) their ability to consider the morphology and the material of BHAs when designing the controllers, (ii) their core mechanisms are based on CPPNs, whose properties help to produce control patterns that are significant in the emergence of efficient movement \cite{cheney2014unshackling}.
 
Although the difference is minimal when the performance of NEAT and HyperNEAT are compared, NEAT demonstrates a more suitable performance. Arguably, two factors affected the performance of HyperNEAT: (a) the substrate design implemented for experimentation, and (b) the absence of geometrical aspects of the domain problem that could not be embodied in the design of the substrate. Furthermore, NEAT was able to design more compact and, hence, more efficient controllers than HyperNEAT, due to the fact that the search space included the number of neurons, their connections and their activation functions. In contrast, the search space, where HyperNEAT operated, is more restricted since it only included the connections in a fixed number of neurons with the same activation function.

Future work directions considering the results gathered from this research can explore the suitability of the NE-based approaches in more realistic scenarios where elements of the environment (e.g., viscosity and friction) are included during simulations. Furthermore, adding periodic activation functions, such as tangent and cosine, could induce other patterns of more efficient upward bending movements. Another avenue of future work may focus on improving the performance of HyperNEAT with a broader exploration of hyper-parameters, such as the number of neurons, hidden layers, and activation functions used to design the substrate.



\section*{{Acknowledgements}}

This project has received funding from the European Union’s Horizon Europe research and innovation programme under grant agreement No. 101070328. UWE researchers were funded by the UK Research and Innovation grant No. 10044516.

\section*{Disclosure of Interests.}
The authors have no competing interests to declare that are relevant to the content of this article.

\end{document}